\documentclass[10pt,twocolumn,letterpaper]{article}

\usepackage{cvpr}
\usepackage{times}
\usepackage{epsfig}
\usepackage{graphicx}
\usepackage{amsmath}
\usepackage{amssymb}
\usepackage{subcaption}
\usepackage{multirow}
\usepackage{cite}
\usepackage{ctable}
\usepackage{comment}

\usepackage[pagebackref=true,breaklinks=true,letterpaper=true,colorlinks,bookmarks=false]{hyperref}
\newcommand*{\rom}[1]{\expandafter\@slowromancap\romannumeral #1@}
\cvprfinalcopy

\begin{document}

\title{(De)Constructing Bias on Skin Lesion Datasets}
\author{Alceu Bissoto\textsuperscript{1} ~~~Michel Fornaciali\textsuperscript{2} ~~~Eduardo Valle\textsuperscript{2} ~~~Sandra Avila\textsuperscript{1}\\
\textsuperscript{1}Institute of Computing (IC) ~~~\textsuperscript{2}School of Electrical and Computing Engineering (FEEC)\\ 
RECOD Lab., University of Campinas (UNICAMP), Brazil\\
}

\maketitle
\thispagestyle{empty}
\begin{abstract}
Melanoma is the deadliest form of skin cancer. Automated skin lesion analysis plays an important role for early detection. Nowadays, the ISIC Archive and the Atlas of Dermoscopy dataset are the most employed skin lesion sources to benchmark deep-learning based tools. However, all datasets contain biases, often unintentional, due to how they were acquired and annotated. Those biases distort the performance of machine-learning models, creating spurious correlations that the models can unfairly exploit, or, contrarily destroying cogent correlations that the models could learn. In this paper, we propose a~set of experiments that reveal both types of biases, positive and negative, in existing skin lesion datasets. Our results show that models can correctly classify skin lesion images without clinically-meaningful information: disturbingly, the machine-learning model learned over images where no information about the lesion remains, presents an accuracy above the AI benchmark curated with dermatologists' performances. That strongly suggests spurious correlations guiding the models. We fed models with additional clinically meaningful information, which failed to improve the results even slightly, suggesting the destruction of cogent correlations. Our main findings raise awareness of the limitations of models trained and evaluated in small datasets such as the ones we evaluated, and may suggest future guidelines for models intended for real-world deployment.

\end{abstract}

\section{Introduction}\label{sec:intro}

The amount of people diagnosed with melanoma is rapidly increasing in the past decades. Today, it is already treated as a public health challenge, especially in high sun exposition areas with Caucasian populations\footnote{\url{http://www.cancer.net/cancer-types/melanoma/statistics}}. Melanoma is the deadliest form of skin cancer, and early detection is crucial~\cite{survival-rates-skin-cancer} for good prognosis, creating a need for efficient early-detection techniques, and thus an incentive for research on automated detection.     

Deep learning methods are the state-of-the-art on skin cancer classification~\cite{esteva2017,haenssle2018man}. That task is challenging due to the vast visual variability of skin lesions, and the subtlety of the cues that differentiate benign and malignant cases. To compound the difficulty, datasets to train the data-hungry models are small, when compared with general-purpose image datasets (\eg, ImageNet, MSCOCO, LabelMe).

Due to the scarcity of good-quality, annotated skin lesion images, two datasets dominate research on automated skin lesion analysis: the Interactive Atlas of Dermoscopy~\cite{argenziano2002dermoscopy} and the ISIC Archive~\cite{isicarchive}. The Atlas is an educational medical resource, with many standardized metadata over the cases it contains, while the ISIC Archive is a much larger, but also less controlled dataset, with images of different sources. Nowadays nearly every reproducible work in the field refer to these datasets for training, evaluating or comparing its models \cite{celebi2015state, valle2018data, bissoto2018skin, brinker2019comparing}, and the ISIC Archive deserves special mention as the source of the images used in the ISIC Challenge~\cite{isic2016,isic2017,isic2018}, an annual event where different teams compare the performance of their algorithms under the controlled supervision of the organizers. 

The problem of having so few, relatively small datasets dominating much of research in automated skin analysis, is the risk of datasets biases. Indeed, the (re)use of relatively small datasets by a research community poses certain risks for research on Machine Learning \cite{peng2011reproducible}. Dataset biases may both inflate the performance of models (presenting them features that are not truthful to real-world data), or play down their performance (by destroying correlations that occur in real-world data, and thus preventing models from exploiting them).
If we think of general datasets, there can be bias over the scenes (rural or urban), acquisition methods (professional or amateur), amount of objects in the scene, angles of views, among other factors~\cite{torralba2011unbiased}. 

If bias is present even in bigger and more diverse datasets~\cite{torralba2011unbiased} like ImageNet \cite{ILSVRC15}, it is naive to think it is not present in the smaller and harder to obtain skin cancer datasets, where we lack works identifying the possible sources of dataset bias. We know, however, that there are visible artifacts introduced during the image acquisition process (\eg, dark corners, marker ink, gel bubbles, color charts, ruler marks, skin hair)~\cite{mishra2016overview} that could inflate models performances due to spurious correlations. 

Despite being impossible to wholly eliminate, it is important to understand bias and its sources to further improve our image acquisition processes and deep learning models.

A useful way to measure the first possible effect of a dataset bias (undue inflation of performances due to spurious correlations in the dataset), is a counterfactual experiment, which destroys the cogent information in the data, and measures how much the performance of models drops. Therefore, our first set of experiments follows that procedure, gradually removing information from skin lesion images, and measuring the network performance.
We perform single- (training and testing on the same dataset) and cross-dataset (training on ISIC and testing on Atlas) experiments, and find that in both cases, the networks are able to maintain a surprising amount of accuracy, even after almost all cogent information has been destroyed. 

Measuring the second possible effect (inability to provide useful correlations for learning) is much harder, since we cannot, \textit{a priori} prove those correlations exist in the real-world, neither that the machine-learning model would learn from them if they were correctly represented in the dataset. The best we can do is to provide additional evidence for the models that we expect would be useful for a human, and measure if that makes any difference.

Thus, in our second experiment set, we add progressively more features, based upon fine-grained dermoscopic attributes (pigment network, negative network, streaks, milia-like cysts, and globules) spatially located on the lesions. In order to provide those features, we employ the annotations available for the Task 2 (Lesion Attribute Detection) of the ISIC Challenge. We expected that such clinically-meaningful skin lesion information would improve the network learning process, but in fact, the performance fails to improve in all scenarios we tested, even when we feed the network with all the image's pixels with an additional channel containing extra clinically-meaningful information.

Summarizing, the main contributions in this work are: 
\begin{itemize}
    \item We assess whether the models' performance are inflated due to dataset bias by performing a counterfactual experiment, where we gradually destroy meaningful information in the data, and measure the performance of our models.
    \item We evaluate whether the dataset is providing useful correlations for learning, by gradually feeding the network with extra clinically-meaningful information.
    \item We provide a discussion to raise awareness of bias in the automated skin lesion analysis community to improve the next generation of solutions for classifying skin lesions in the real world.
\end{itemize}

We organized the text as follows. We introduce our motivation and related works in Section \ref{sec:intro}. We present our methodology, materials and goals in Section \ref{sec:matmeth}. We detail our experiment to gradually destroy clinically-meaningful information in skin lesion images and evaluate the network's responses in Section \ref{sec:destruct}. We detail our experiment where we try to guide the network's learning process through additional clinical information in Section \ref{sec:construct}. Finally, we review and discuss our findings in Section \ref{sec:conclusion}. 

\section{Materials and Methods}\label{sec:matmeth}

\subsection{Datasets}

We employ two of the most important skin lesion datasets: the Interactive Atlas of Dermoscopy (Atlas)~\cite{argenziano2002dermoscopy} and the International Skin Imaging Collaboration (ISIC) Archive~\cite{isicarchive}. 
While Atlas excels for having rich metadata associated with each lesion image, ISIC excels for being diverse. Since both are publicly available, most of the recent works in skin lesion analysis rely on these datasets. When exploring reproducible works on lesion segmentation \cite{celebi2015state, xue2018adversarial}, dermoscopic attribute segmentation \cite{kawahara2018fully}, skin lesion classification \cite{valle2018data, brinker2019comparing, perez2018data}, or skin lesion synthesis \cite{bissoto2018skin} those two datasets are almost certain to be included.
Next we describe their individual characteristics, and discuss how they differ.

The \textbf{Atlas}~\cite{argenziano2002dermoscopy} is a medical educational dataset composed of $+1,000$ cases of pigmented skin lesions. Each case is associated with clinical and dermoscopic images. Each skin lesion has clinical data (\eg, location, diameter, elevation), histopathological results, diagnosis, and the presence or absence of dermoscopic attributes. The presence of those rich metadata correspond to the pedagogical objectives of the Atlas of teaching dermoscopy through reliable and understandable medical algorithms (\eg, the 7-point checklist). The Atlas also groups the lesions according to their level of diagnostic difficulty (low, medium or high), which indicates how difficult it is to identify the medical attributes (\eg, networks, dots-and-globules, etc.) in the lesions. The difficulty relies on the morphological variability of a given criterion, which explains the sometimes low intra- and interobserver agreement of such medical algorithm.

Since it is a dataset for medical education purposes, the statistics of its metadata do not necessarily reflect their occurrences in any real-world population. In this work, we are especially interested in the dermoscopic attributes annotation. Lesions' dermoscopic attributes analysis (through pattern-based medical algorithms) is crucial for dermatologists to diagnose skin cancer. This information enable us to verify bias by comparing the medical algorithm performance, the network performance, and an Artificial Intelligence benchmark for melanoma classification~\cite{brinker2019comparing}. 
Segmentation masks, which is also relevant to this discussion are not available, but we employ computational methods to obtain them.

For our experiments, we select only the dermoscopic samples, remove ``duplicates'' (some medical cases have multiple images), and include only the classes present in the dataset of task~$2$ of $2018$ ISIC Challenge (melanoma, nevus, and seborrheic keratosis). Those alterations result in a dataset containing $872$ images.  

The \textbf{ISIC Archive (ISIC)} dataset~\cite{isicarchive} is a bigger and more generic dataset, composed of more than $23,000$ images collected from different leading clinical centers internationally, using a variety of devices for acquisition. Since the first ISIC Challenge in 2016 \cite{isic2016}, this dataset is increasing in size and in the amount of information available for each lesion. Segmentation masks and maps over five dermoscopic attributes (pigment network, negative network, streaks, milia-like cysts, and globules) are available for smaller subsets of the dataset. 

It is import to note that the dermoscopic attributes annotations in ISIC and Atlas differ in two ways. First, in ISIC the annotation is a mask that maps the dermoscopic attributes in the original images. In the Atlas dataset, we only have the information about the presence or absence of each dermoscopic attribute. Second, the two datasets annotated information about different dermoscopic attributes, with different levels of detail. Unfortunately, only the patterns present in the Atlas dataset allow to apply (and evaluate) the medical pattern-based algorithms . 

For all of this work's experiments, we use only the data from the second task of the ISIC 2018 Challenge \cite{isic2018} for dermoscopic attribute detection. This subset contain $2,594$ lesions' dermoscopic attributes information. 

For \textbf{both} datasets, the class frequencies (types of skin lesions, \eg, melanomas, nevi, keratoses) do not reflect any real-world population. That, however, is a necessity for training and evaluating machine-learning models, since in real-world populations, the proportion of melanomas to nevi, for example, is \textit{extremely} small, generating huge imbalances that most models would not tolerate. The need to ``rebalance'' the classes for machine-learning, however, can generate models biased towards some classes, and inflate the rate of false positive for melanomas, for example.

\subsection{Methodology}

\begin{figure*}[t]
\centering
\begin{subfigure}[b]{\linewidth}
\centering
\includegraphics[width=0.17\linewidth,height=0.126\linewidth]{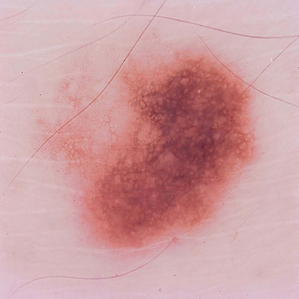}\hspace{0.05cm}
\includegraphics[width=0.17\linewidth,height=0.126\linewidth]{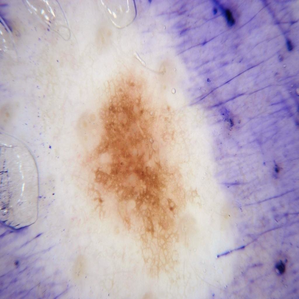}\hspace{0.05cm}
\includegraphics[width=0.17\linewidth,height=0.126\linewidth]{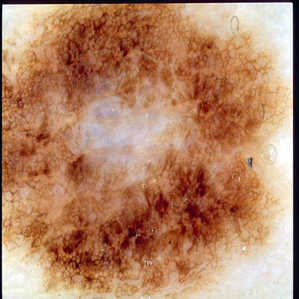}\hspace{0.05cm}
\includegraphics[width=0.17\linewidth,height=0.126\linewidth]{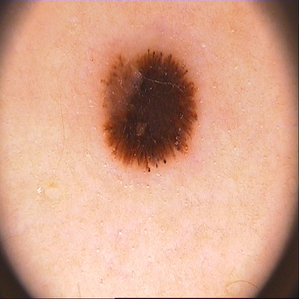}\hspace{0.05cm}
\includegraphics[width=0.17\linewidth,height=0.126\linewidth]{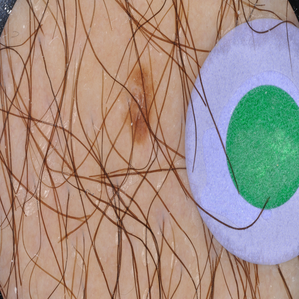}
\caption{Traditional images}\vspace{.05cm}\label{fig:atlasrgb}
\end{subfigure}

\begin{subfigure}[b]{\linewidth}
\centering
\includegraphics[width=0.17\linewidth,height=0.126\linewidth]{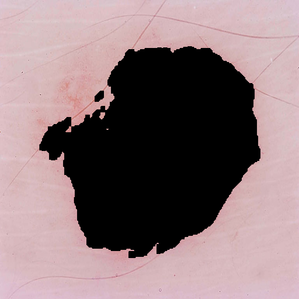}\hspace{0.05cm}
\includegraphics[width=0.17\linewidth,height=0.126\linewidth]{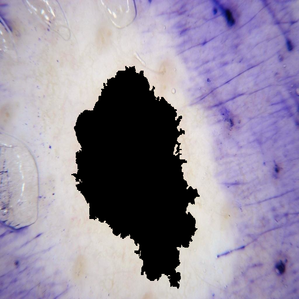}\hspace{0.05cm}
\includegraphics[width=0.17\linewidth,height=0.126\linewidth]{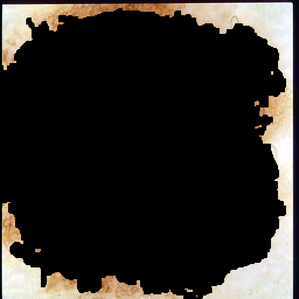}\hspace{0.05cm}
\includegraphics[width=0.17\linewidth,height=0.126\linewidth]{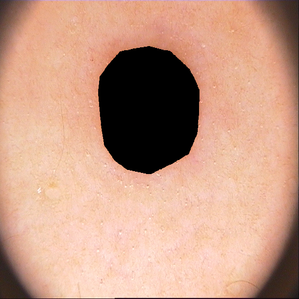}\hspace{0.05cm}
\includegraphics[width=0.17\linewidth,height=0.126\linewidth]{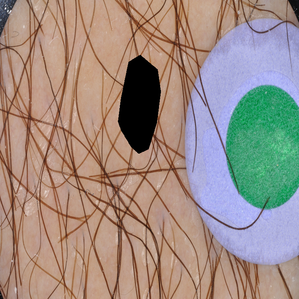}
\caption{Only Skin images}\vspace{.05cm}\label{fig:atlasbg}
\end{subfigure}

\begin{subfigure}[b]{\linewidth}
\centering
\includegraphics[width=0.17\linewidth,height=0.126\linewidth]{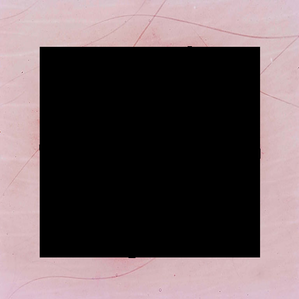}\hspace{0.05cm}
\includegraphics[width=0.17\linewidth,height=0.126\linewidth]{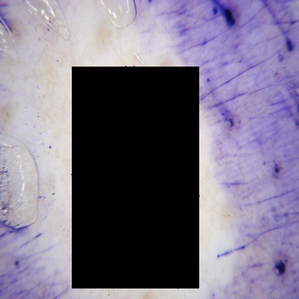}\hspace{0.05cm}
\includegraphics[width=0.17\linewidth,height=0.126\linewidth]{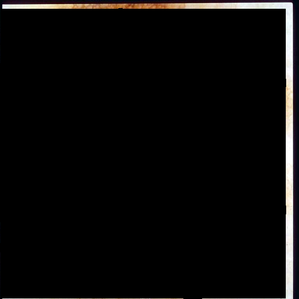}\hspace{0.05cm}
\includegraphics[width=0.17\linewidth,height=0.126\linewidth]{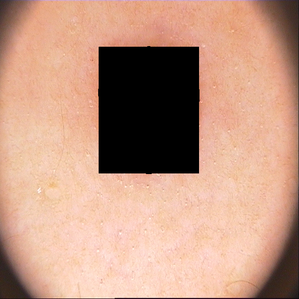}\hspace{0.05cm}
\includegraphics[width=0.17\linewidth,height=0.126\linewidth]{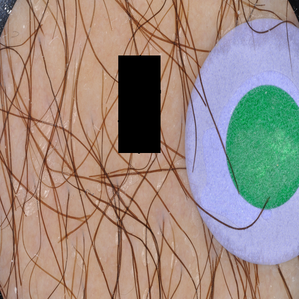}
\caption{Bbox images}\vspace{.05cm}\label{fig:atlasbbox}
\end{subfigure}

\begin{subfigure}[b]{\linewidth}
\centering
\includegraphics[width=0.17\linewidth,height=0.126\linewidth]{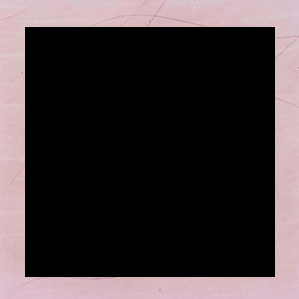}\hspace{0.05cm}
\includegraphics[width=0.17\linewidth,height=0.126\linewidth]{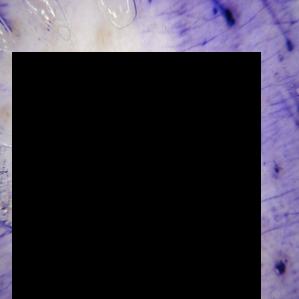}\hspace{0.05cm}
\includegraphics[width=0.17\linewidth,height=0.126\linewidth]{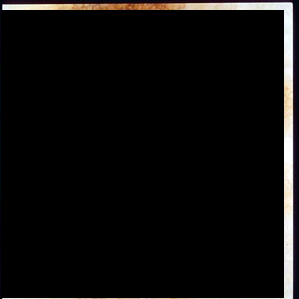}\hspace{0.05cm}
\includegraphics[width=0.17\linewidth,height=0.126\linewidth]{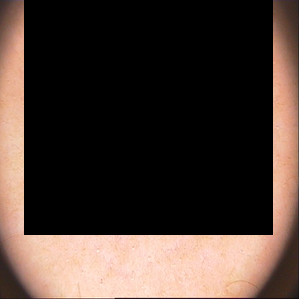}\hspace{0.05cm}
\includegraphics[width=0.17\linewidth,height=0.126\linewidth]{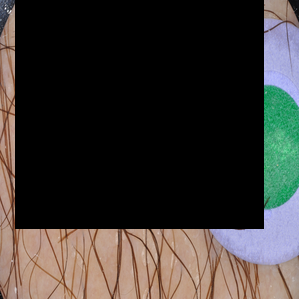}
\caption{Bbox70 images\vspace{-0.1cm}}\label{fig:atlasbbox70}
\end{subfigure}

\caption{Samples from each of our disrupted datasets. We gradually remove cogent information, until there is no information left to apply any aspect of medical score algorithms \cite{7points, friedman1985early}. Next, we use those sets to evaluate if the network can still learn patterns with the information left to correctly classify skin lesions. Best seen in digital format.} \label{fig:atlasmodifications}
\end{figure*}

To evaluate the presence and effect of dataset bias in Atlas and ISIC, we propose to:
\begin{itemize}
    \item Perform destructive actions (see Figure~\ref{fig:atlasmodifications}) in the dataset to analyze if the network can still learn patterns to correctly classify skin lesions, even without clinically-meaningful information available.\vspace{-0.1cm}
    \item Apply the 7-point checklist algorithm~\cite{7points} to the Atlas dataset, and analyze the result comparing it with the recent melanoma classification benchmark for AI~\cite{brinker2019comparing} to verify how biased it is due to its educational purposes and acquisition methods.\vspace{-0.1cm}
    \item Perform constructive actions (see Figure~\ref{fig:isicmodified}) in the dataset, building from clinically-meaningful information to guide the network's learning, to analyze if the result improves.
\end{itemize}

To accomplish our goals, we propose destructive and constructive actions in the target datasets. We present the details of each destructive action and the result data in Section \ref{sec:destruct}, and the details of each constructive action and the result data in Section \ref{sec:construct}. 
Next, we introduce our ideas to exploit the deep neural network learning capabilities. 

\textbf{Destructing Atlas-dataset}: We employ the Atlas dataset with our disruptive actions for both training and testing the network in the destruction of information approach.
We use $10$ splits that we keep the same throughout all sets of images (\textit{Traditional, Only Skin, Bounding Box, Bounding Box 70\%}) to make comparisons fair. To compose each training split, we randomly select $70\%$ of the images of each diagnostic difficulty present in the Atlas dataset (low, medium and high). We compose the corresponding test split using the $30\%$ that is left. Following this procedure, we reduce the possibility of biasing our results with a split that is especially good for a given set of images.
Since the training and test sets come from the same data distribution (same dataset), we expect these results to be optimistic, and that motivates our three next designs.

\textbf{Destructing ISIC-dataset}: We also apply the destruction of information approach to the ISIC dataset.
We do that to confirm the behavior verified in Atlas in a more generic dataset, with fewer effects of human bias. 
We apply the same $10$ split generation procedure we described for this experiment, except for the diagnostic difficulty stratification (the information is not present for the ISIC dataset).

\textbf{Destructing Cross-dataset}: We increase the difficulty by experimenting with a cross-dataset fashion. We train with all $2,594$ samples from the ISIC dataset and evaluate on the complete $872$ images set from Atlas. The differences between the statistics between those two datasets make this task harder, and better reflect a real-world setting \cite{torralba2011unbiased}. We repeat that experiment $10$ times, for statistical significance. 

\textbf{Constructing ISIC-dataset}: We attempt to guide the network's learning using the dermoscopic attribute information available for the ISIC dataset. We create three sets of images (\eg, \textit{Grayscale Attributes, RGB Attributes, Traditional $+$ Grayscale Attributes}), where the amount of information is gradually increased (see Section \ref{sec:construct}). We keep the same training procedure and splits from \textit{Destructing ISIC-dataset}.

\subsection{Training and Evaluation Setup}

We use the same network architecture and hyperparameters for all experiments.
We employ an Inceptionv4 network \cite{szegedy2017inception}, widely used for computer vision, and well-established for skin lesion analysis. 
To train each network, we use Stochastic Gradient Descent (SGD) with momentum $0.9$, weight decay $0.001$ and learning rate $1$e-$3$, which we reduce to $1$e-$4$ after epoch $25$. We use a batch size of $32$, shuffling the data before each epoch.

We fine-tune the ImageNet \cite{ILSVRC15} pre-trained network to the target dataset. We resize the input images to $299\times299$ to fit the input size of Inceptionv4. To augment the dataset \cite{perez2018data}, we apply random horizontal and vertical flips, random resized crops that contain from $75\%$ to $100\%$ of the original image, random rotations between $-45$ and $45$ degrees, and random hue changes between $-20\%$ to $20\%$.
We apply the same augmentations on both train and test. For the evaluation, we average the predictions over $50$ augmented versions of each image. We normalize the input using the z-score, computed on ImageNet's training set mean and standard deviation. For all experiments, we report the Area Under the ROC Curve (AUC). 

We limit our datasets to contain lesions' dermoscopic attributes for every sample, shrinking ISIC considerably. 
Since both our datasets are relatively small, we choose not to use a validation set, using the weights after the $60$th epoch for test evaluation\footnote{All our source code is readily available on \url{https://github.com/alceubissoto/deconstructing-bias-skin-lesion}.}. 

\section{Information Destruction Experiments}\label{sec:destruct}

In this section, we detail our information destruction experiments. We intend to investigate the presence of dataset bias by gradually removing cogent information. First, we introduce the disrupted datasets used and proceed to show and discuss our results.

\subsection{Data}

Next, we present the different datasets modifications made for our first experiments and our motivations behind each one. In Figure~\ref{fig:atlasmodifications} we show examples of each variation. We point that we keep the same modifications for both training and testing our networks.

\textbf{Traditional}: This dataset contains the usual information used for training and evaluating skin lesion analysis networks. The images contain all pixels' information and we expect it to have the highest scores in our tests, being our upper bound baseline. 

\textbf{Only Skin}: To create this dataset, we take advantage of segmentation masks. We apply the mask in the samples from the \textit{Traditional} dataset, removing the pixels' information (they turn black) inside the actual lesion. We keep only the silhouette of the lesion and the skin of the image. Our intention when creating this dataset is to destroy the lesion information while verifying if the network could still make sense of the remained pixels to classify the samples correctly. Unlike the ISIC dataset, the Atlas dataset does not provide the lesions' ground truth segmentation masks. To obtain them, we choose to use the SeGAN model~\cite{xue2018adversarial}, which placed $4$th on the segmentation task at the 2018 ISIC Challenge making use of a generative approach for skin lesion segmentation. 

\textbf{Bounding Box (Bbox)}: The lesion border is an essential feature to diagnose skin lesions. The classic ABCD medical algorithm~\cite{abcd} consider this feature, which accounts border symmetry and border regularity. To destroy this information from the dataset, we cover the silhouette of the lesion with a black bounding box. At this point, we already removed the lesion and its borders information. Only healthy skin and artifacts reminiscent from the image acquisition process are available for the network to learn.

\textbf{Bounding Box $70$\% (Bbox$70$)}: The diameter (size) of the lesion is considered by dermatologists to diagnose skin lesions since melanomas are usually bigger (start with a diameter of more than $6$mm \cite{friedman1985early} than benign lesions. The diameter is the last clinical feature we attempt to remove from the network's learning possibilities. For this purpose, we define that every bounding box must at least have the size of a $250\times250$ square (note that images are $299\times299$). We keep intact bounding boxes that need to be bigger to cover the lesion. The $250\times250$ square is sufficient to cover $70\%$ of the pixels. We place this square at the center of the lesion. If the lesion is not in the center of the image, part of the box is not visible. In these specific cases, it is possible that the bounding boxes cover less than $70\%$ of the pixels.
At this point, there is no information left to apply any of the factors from the ABCD~\cite{friedman1985early, abcd}, ABCDE~\cite{abcde} or any pattern-based algorithm~\cite{7points}.

\subsection{Results and Discussion}

We employ the melanoma classification benchmark \cite{brinker2019comparing} to measure the expected performance for dermatologists, in an unbiased scenario. 
This benchmark is the result of a study with $157$ German dermatologists to be a reliable benchmark for artificial intelligence algorithms. Brinker \etal's procedure were to send an electronic questionnaire to dermatologists containing $100$ dermoscopic images ($80$~nevi and $20$ biopsy-verified melanoma) randomly chosen from the ISIC Archive, asking for their evaluation. The AUC achieved by dermatologists for dermoscopic images (which is the case for our Atlas set) is $67\%$. 

\begin{figure}[t]
    \centering
    \includegraphics[trim={0 0.1cm 0 0.7cm},clip,width=\linewidth]{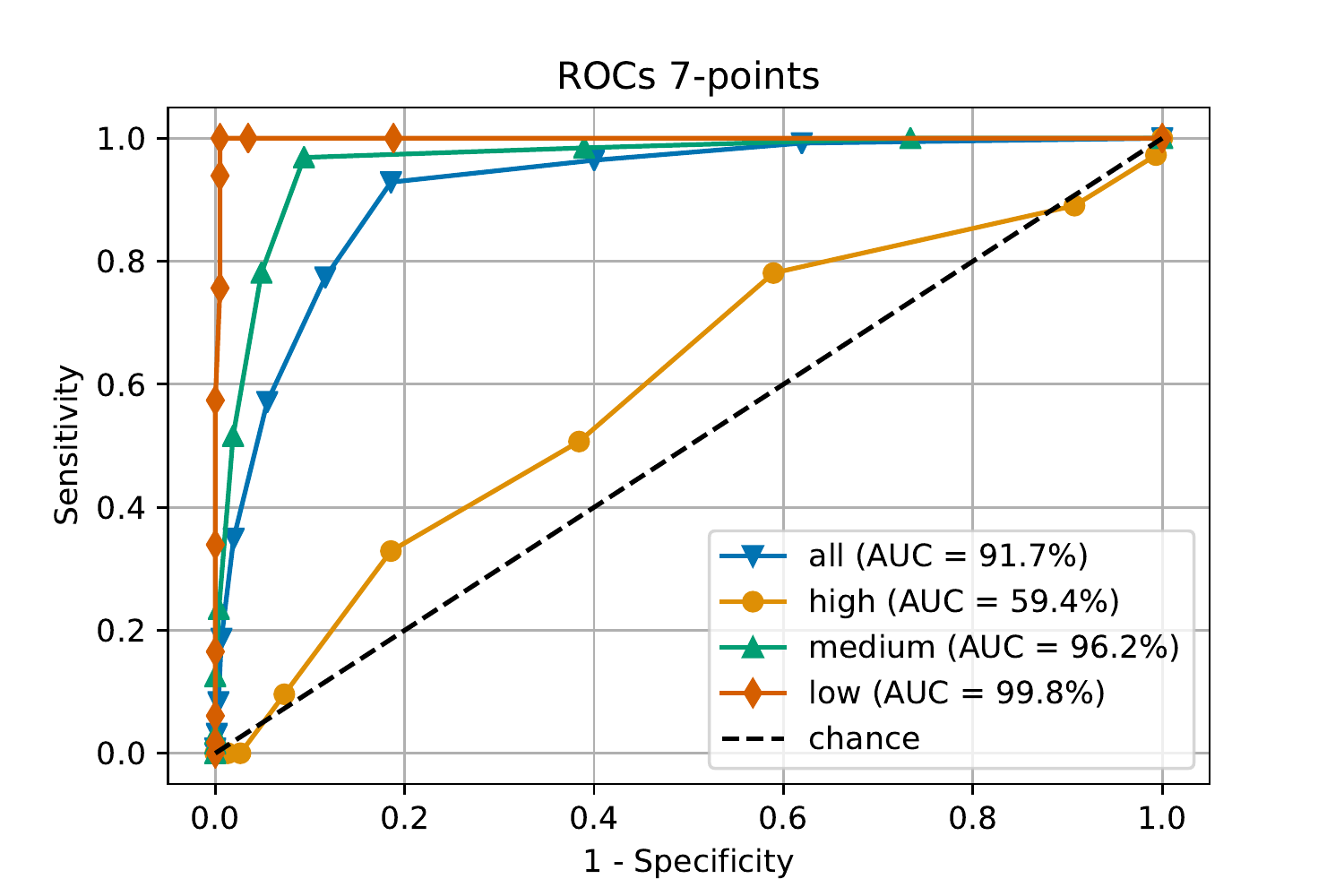}
    \caption{Performance of the 7-point checklist algorithm on the Atlas dataset. It shows a huge gap to the performance of dermatologists evaluated in $100$ random dermoscopic samples from the ISIC Archive, which is $67\%$ \cite{brinker2019comparing}. The results for 7-point checklist applied on Atlas is optimistic considering the dataset's bias towards its educational aspects.}
    \label{fig:7pointsresult}
\end{figure}

We employ 7-point checklist~\cite{7points}, a score-based medical algorithm, to verify bias in the Atlas dataset. This way we can isolate the neural network's learning capabilities. Dermatologists use attribution pattern analysis to diagnose malignant cases. The 7-point medical algorithm assigns a score to each of the dermoscopic attributes. The medical practitioner needs to accumulate the scores over the detected present attributes. If this score surpasses a threshold, the lesion is assigned as a melanoma. Dermatologists use this information in addition to clinical information (if the lesion is growing, if it itches, if it bleeds, if it hurts, its location and patient's age and sex), to diagnose skin lesions. We use the 7-points checklist score available as metadata of the Atlas dataset\footnote{\url{http://derm.cs.sfu.ca}}. It achieves $91.7\%$ AUC over all selected Atlas samples (see Figure~\ref{fig:7pointsresult}). 

The huge gap between the 7-point checklist performance with the melanoma classification benchmark reveals it is biased due to the characteristics and educational objectives of the Atlas dataset. Low and medium difficulty cases selected to compose the dataset are probably hand-picked to be good examples to teach new medical practitioners to identify and classify dermoscopic attributes, while hard cases are exceptions to the pattern-based analysis. 

\begin{figure}
    \centering
    \includegraphics[trim={0 0.55cm 0 1.02cm},clip,width=1.025\linewidth]{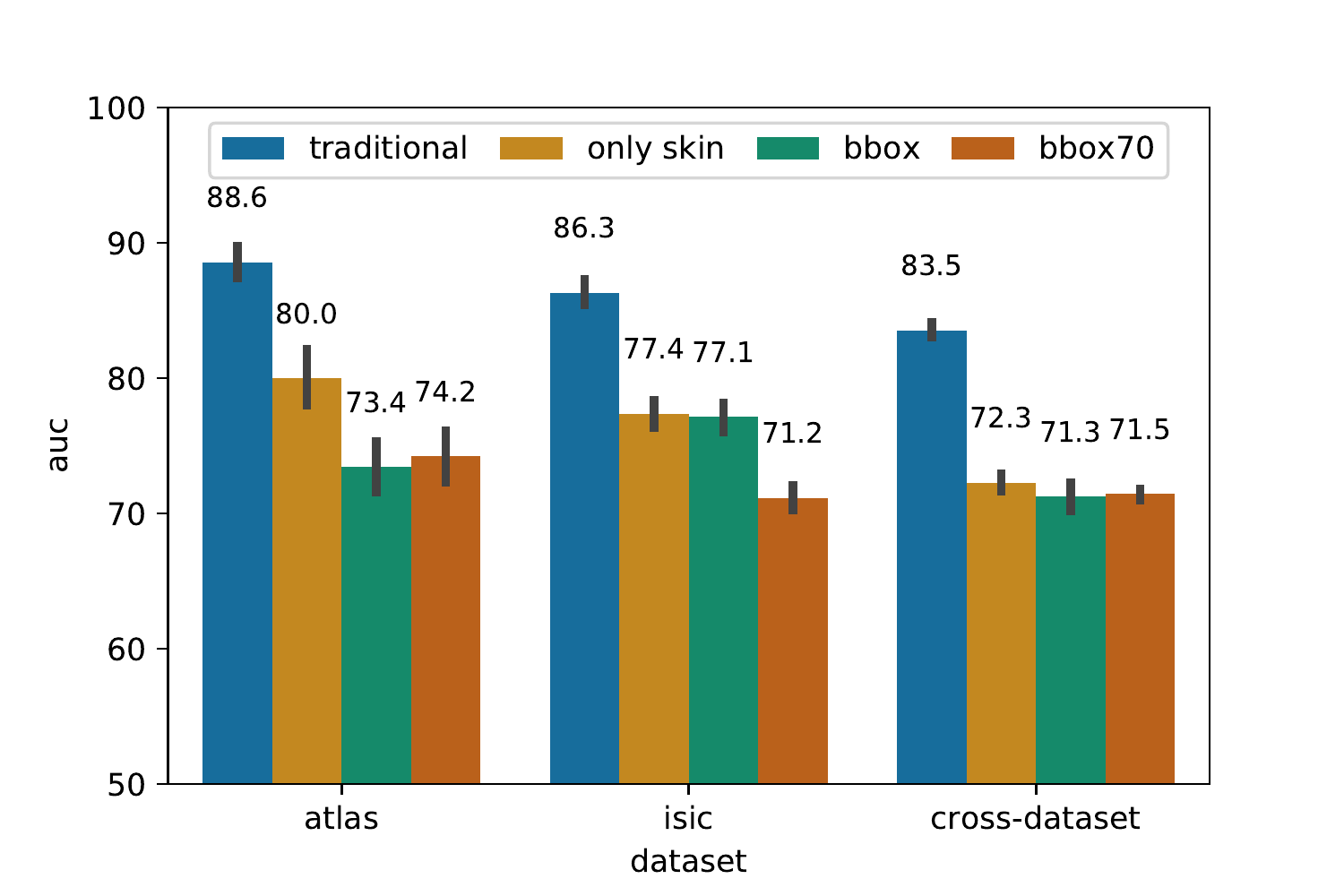}
    \caption{Models' performance over the disturbed datasets. We first remove all the pixel colors inside the lesion (\textit{only skin}), proceeding to remove border information (\textit{bbox}), and finally, removing the size (diameter) of the lesion (\textit{bbox70}). Surprisingly, even when we destruct all clinical-meaningful information, the network finds a way to learn to classify skin lesion images much better than chance.}
    \label{fig:deconstructionall}
\end{figure}

\begin{figure*}
    \centering
    \includegraphics[trim={0 0.42cm 0 0.41cm},clip,width=0.98\linewidth]{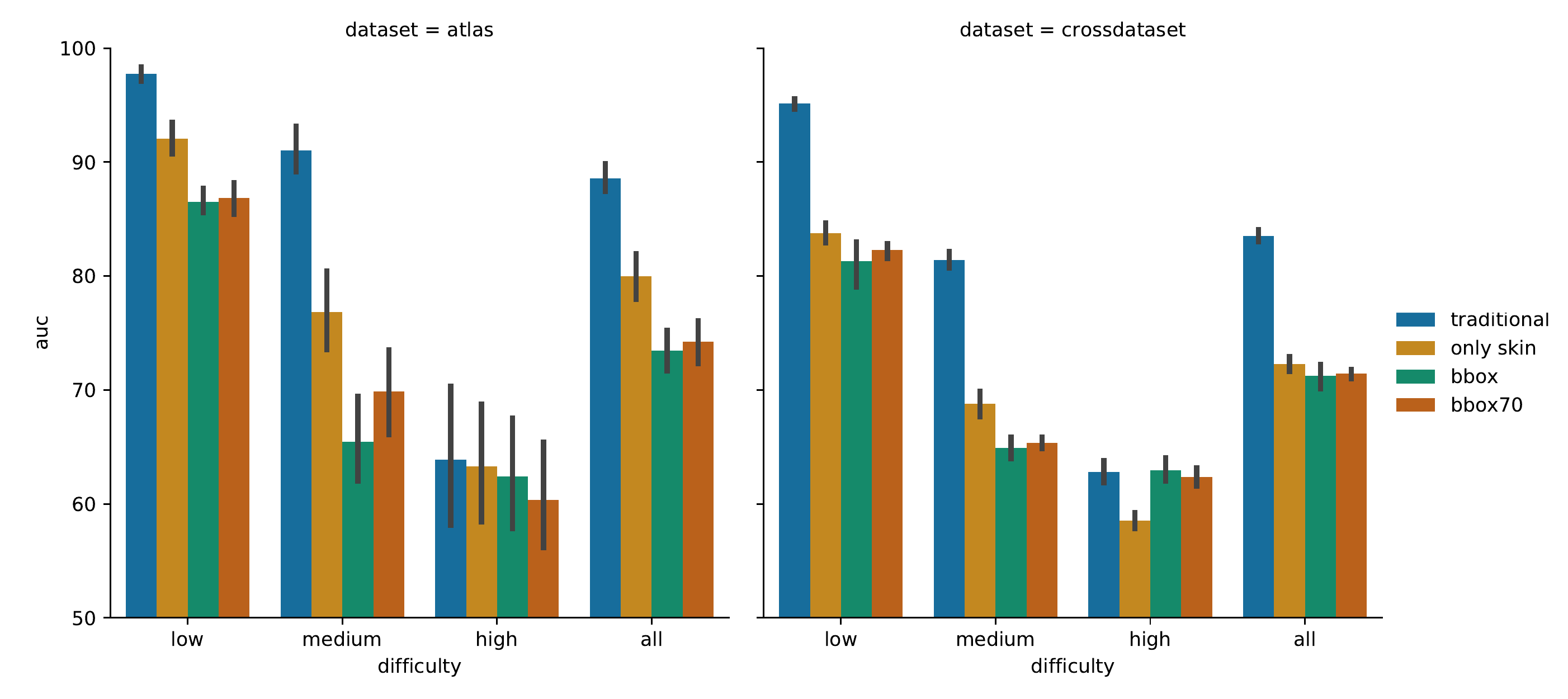}
    \caption{We show the differences over the disturbed datasets, stratifying the performance into the different diagnostic difficulties. High difficulty diagnostic present resilience to the removal of cogent information. Despite not presenting as high numbers as the other difficulties, they are still much better than chance, revealing the patterns learned are not clinical. Other difficulties are more affected by the disturbances, but the overall result for \emph{bbox}, and even \emph{bbox70}, shockingly surpasses melanoma classification benchmark\cite{brinker2019comparing} of $67\%$. This result suggests that dataset bias inflates our model's results.}
    \label{fig:deconstructiondiffs}
\end{figure*}

Next, we try to find the source of bias, by gradually destructing clinical-meaningful information from the images, and assessing the network's performance on them. Figures~\ref{fig:deconstructionall} and \ref{fig:deconstructiondiffs} show the network's performance for the different sets in the Atlas, Cross-dataset, and ISIC experiments respectively. 

High difficulty lesions classification seem to be a very hard and specific task to the network, as it is for dermatologists. It could not learn clinical patterns properly with the training set, and destroying information do not influence the results. 
We understand that the network is probably exploiting image acquisition artifacts and dataset bias.

When experimenting in a cross-dataset fashion, the performance drops as expected, because of the differences between the statistics of Atlas and ISIC. The behavior of the network is similar in all experiments, and the following analysis can be generalized.   

\textit{Traditional} has the best overall performance, as expected. The network results follow the annotation of difficulty to diagnose by dermatologists. 
The results start to drop in \textit{Only Skin}, where we start to deconstruct the information. When we remove the pixel information inside the lesion, we are removing all the information about dermoscopic attributes. The only clinically-meaningful information present is the border of the lesion, that could be used to verify its symmetry and irregularity, and skin features, such as vascularization. 

When we remove the information of the borders, on \textit{Bbox}, the performance lower, even more, revealing that we removed an essential feature for classification. An explanation, referring to medical algorithms like ABCD \cite{abcd}, is that the diameter of the box contains the information on the size of the lesion, which is also relevant information when diagnosing skin lesions.

At \textit{Bbox$70$}, we remove $70\%$ of all pixels in the image and all medical relevant features that could aid the classification. Still, surprisingly, the network can make sense of visual features to make decisions that are much better than chance. There is a pattern within the available pixels that contain information that leads to the correct label. This is shocking. The numbers achieved by the network at this point even surpass the AUC achieved by dermatologists on the melanoma classification benchmark.
As sanity check, we performed an experiment hiding all image information, feeding the network (for training and testing) only zero-filled images. We achieved an AUC of 50\%, which is expected since AUC is insensitive to class balance. 

We believe that dataset bias is the culprit for inflating the network's performance in our destructive experiments, introducing artifacts \cite{mishra2016overview} that undesirably can deviate the network's attention from more critical features. We also verify that bias is not only present in the smaller educational purpose Atlas dataset, but also the most diversified ISIC dataset. Even performing the experiments in a cross-dataset fashion (the network is trained on ISIC, and tested on Atlas), the unnatural behavior persists, attesting to the fact that these two datasets may also share the same bias. We will address the exact causes and artifacts in future works.

Another possibility is that there is meaningful information at the borders of the images (parts that were not affected by the destruction procedures). This is unlikely because according to medical algorithms \cite{7points, abcd, abcde}, there is no information left to account.

\section{Information Construction Experiments}\label{sec:construct}

Since we have masks that maps the dermoscopic attributes in the lesion, we want to verify if we can simplify and guide the learning process by feeding the network with that detailed clinically-meaningful information. We gradually increase the amount of information fed to the network, building from only the attributes information. We describe each set of data in the next subsection. 

\subsection{Data}

\begin{figure*}[t]
\centering

\begin{subfigure}[b]{\linewidth}
\centering
\includegraphics[width=0.17\linewidth,height=0.126\linewidth]{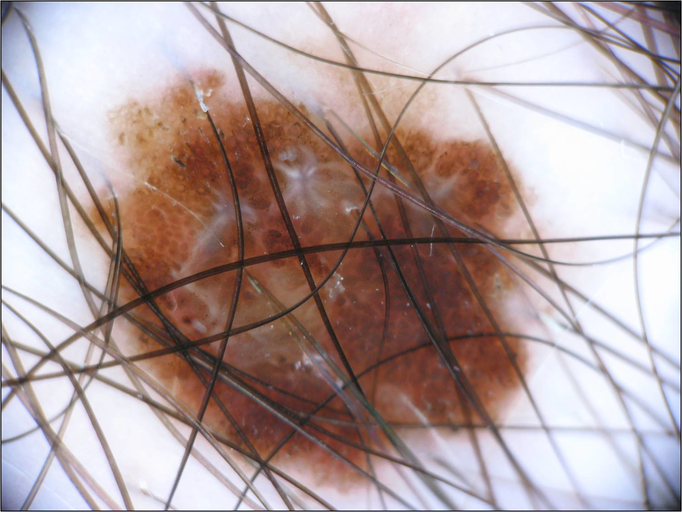}\hspace{0.05cm}
\includegraphics[width=0.17\linewidth,height=0.126\linewidth]{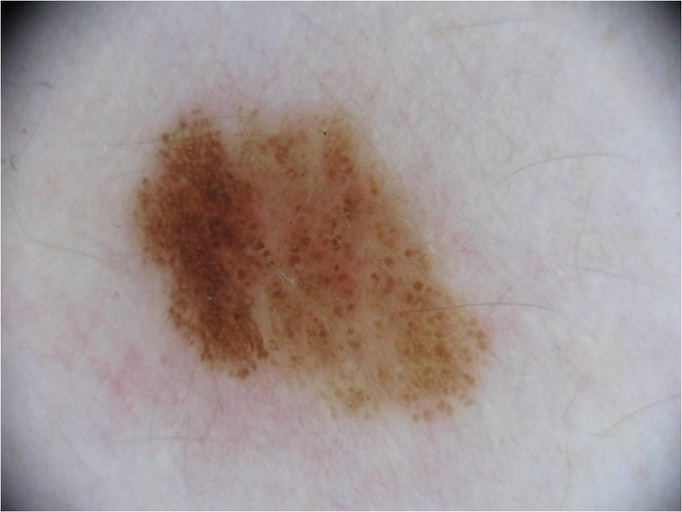}\hspace{0.05cm}
\includegraphics[width=0.17\linewidth,height=0.126\linewidth]{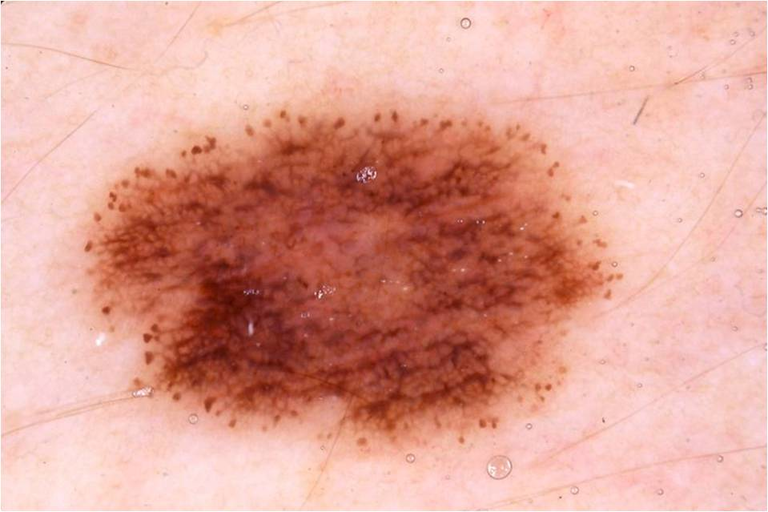}\hspace{0.05cm}
\includegraphics[width=0.17\linewidth,height=0.126\linewidth]{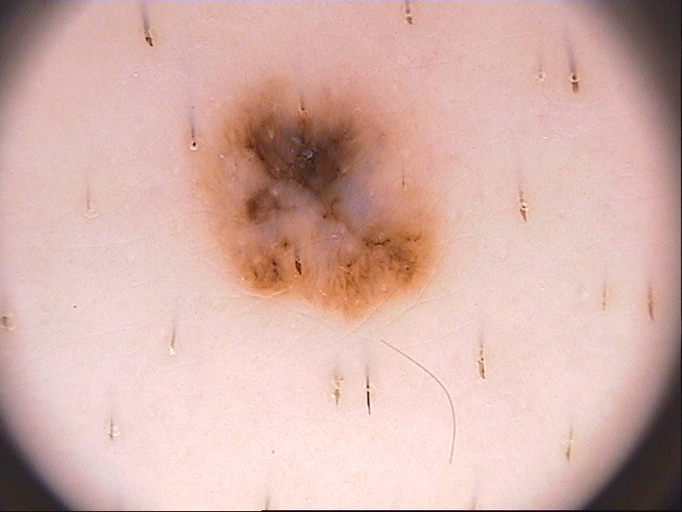}\hspace{0.05cm}
\includegraphics[width=0.17\linewidth,height=0.126\linewidth]{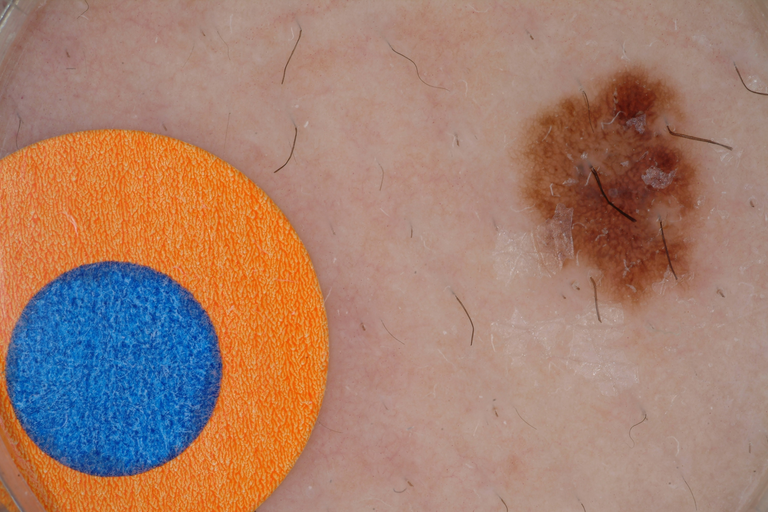}
\caption{Traditional images}\vspace{.05cm}
\end{subfigure}

\begin{subfigure}[b]{\linewidth}
\centering
\includegraphics[width=0.17\linewidth,height=0.126\linewidth]{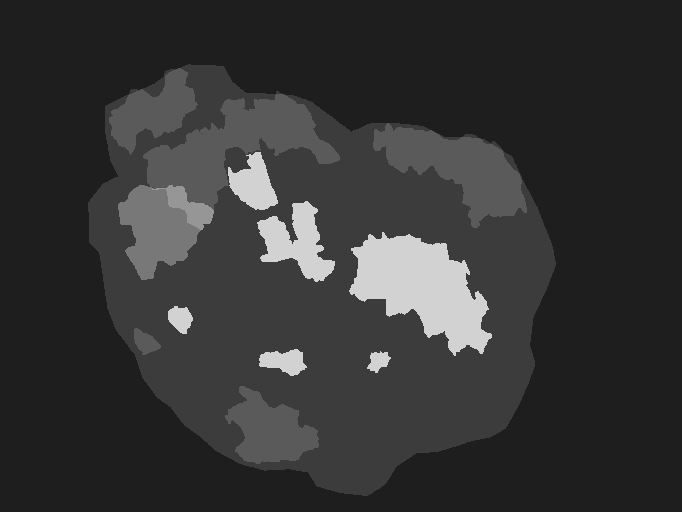}\hspace{0.05cm}
\includegraphics[width=0.17\linewidth,height=0.126\linewidth]{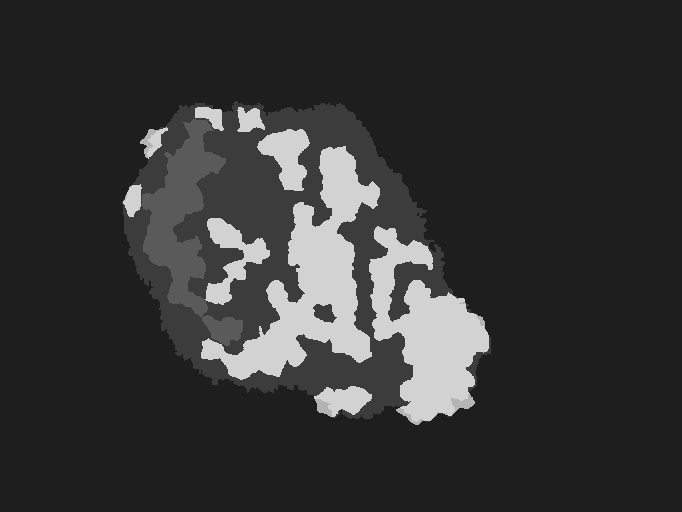}\hspace{0.05cm}
\includegraphics[width=0.17\linewidth,height=0.126\linewidth]{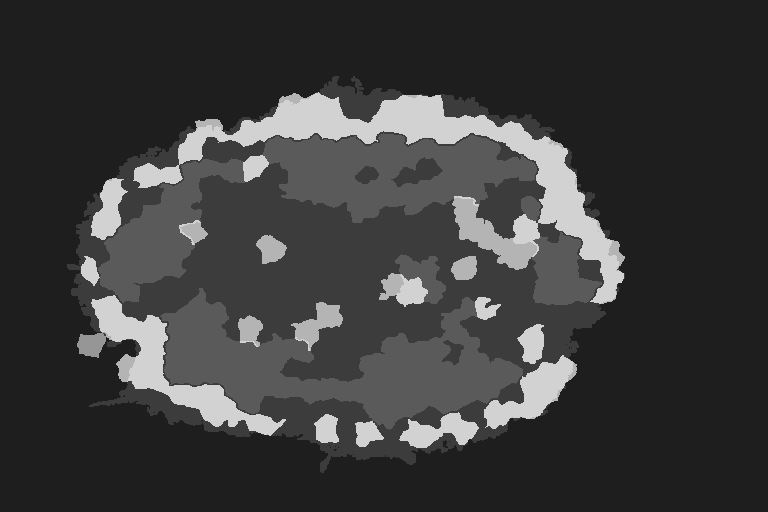}\hspace{0.05cm}
\includegraphics[width=0.17\linewidth,height=0.126\linewidth]{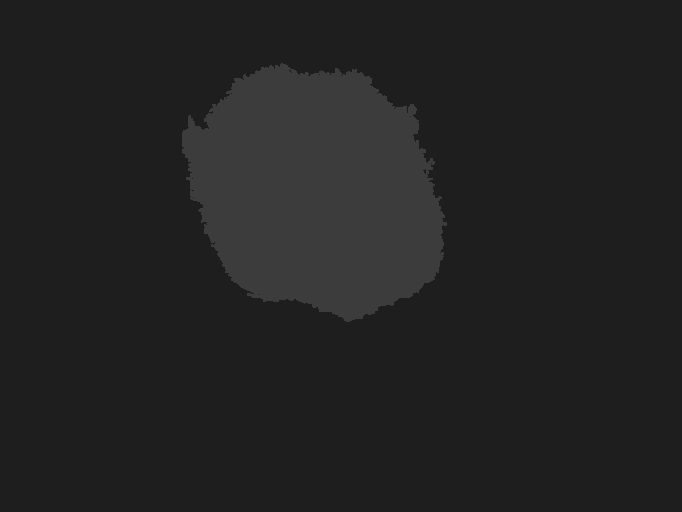}\hspace{0.05cm}
\includegraphics[width=0.17\linewidth,height=0.126\linewidth]{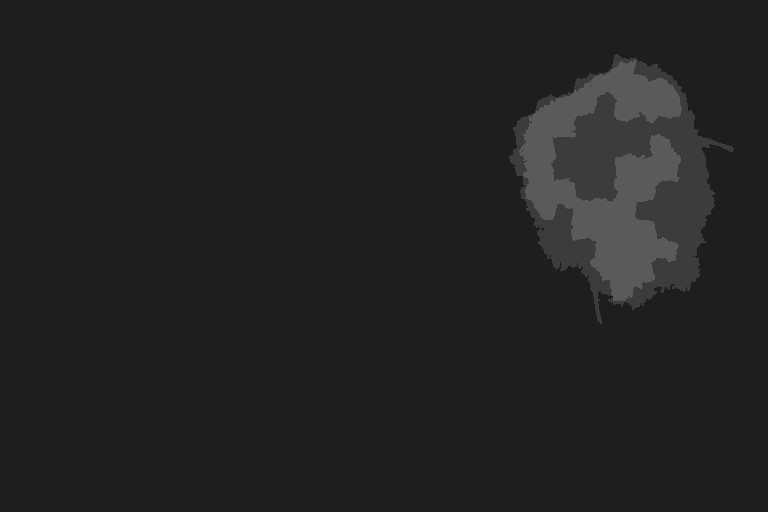}
\caption{Grayscale Attributes images}\vspace{.05cm}
\end{subfigure}

\begin{subfigure}[b]{\linewidth}
\centering
\includegraphics[width=0.17\linewidth,height=0.126\linewidth]{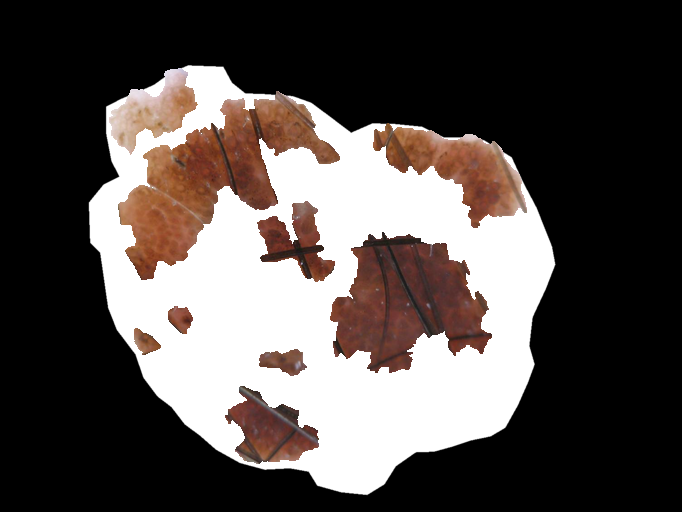}\hspace{0.05cm}
\includegraphics[width=0.17\linewidth,height=0.126\linewidth]{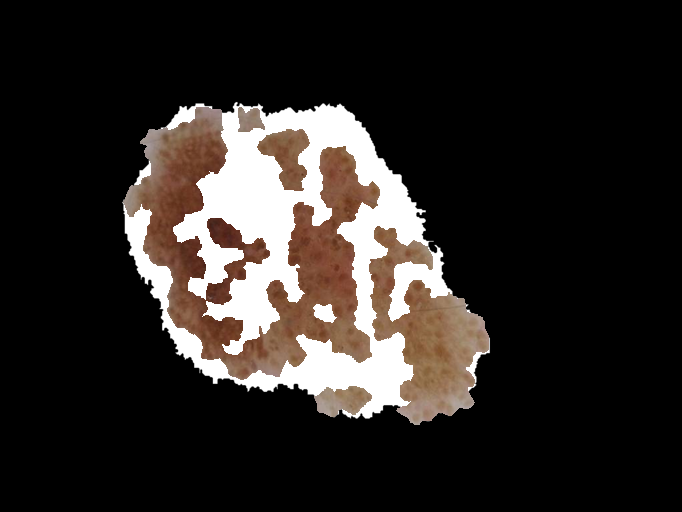}\hspace{0.05cm}
\includegraphics[width=0.17\linewidth,height=0.126\linewidth]{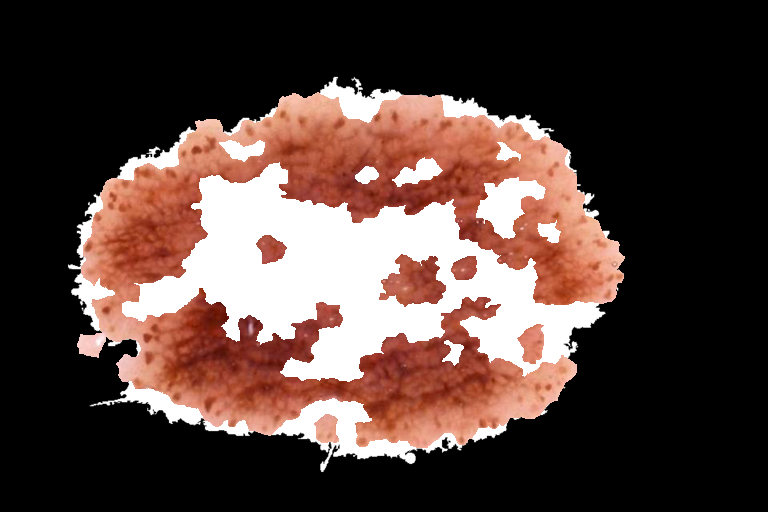}\hspace{0.05cm}
\includegraphics[width=0.17\linewidth,height=0.126\linewidth]{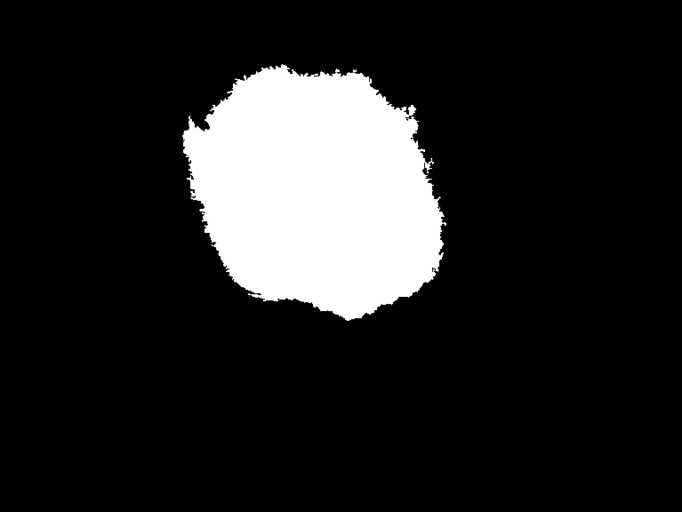}\hspace{0.05cm}
\includegraphics[width=0.17\linewidth,height=0.126\linewidth]{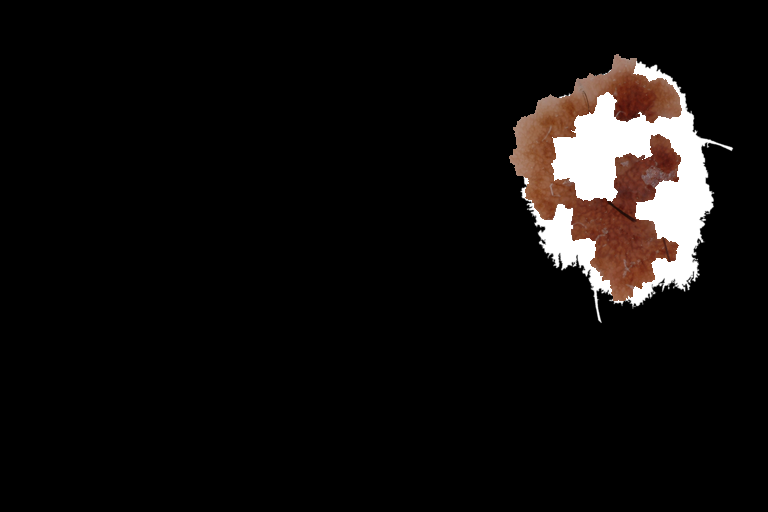}
\caption{RGB Attributes images}\vspace{.05cm}
\end{subfigure}

\begin{subfigure}[b]{\linewidth}
\centering
\includegraphics[width=0.17\linewidth,height=0.126\linewidth]{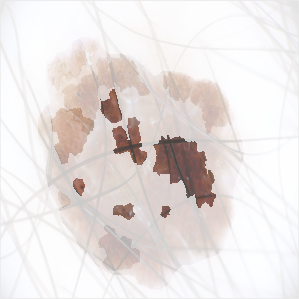}\hspace{0.05cm}
\includegraphics[width=0.17\linewidth,height=0.126\linewidth]{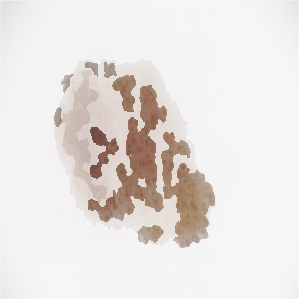}\hspace{0.05cm}
\includegraphics[width=0.17\linewidth,height=0.126\linewidth]{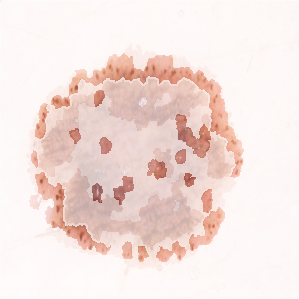}\hspace{0.05cm}
\includegraphics[width=0.17\linewidth,height=0.126\linewidth]{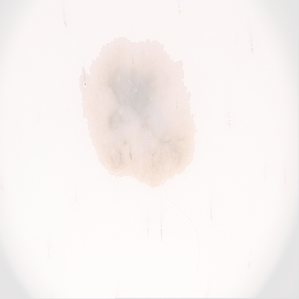}\hspace{0.05cm}
\includegraphics[width=0.17\linewidth,height=0.126\linewidth]{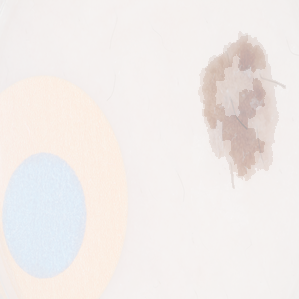}
\caption{Traditional$+$Grayscale Attributes images}
\end{subfigure}

\caption{Samples from each of the variations created for the information construction experiment. We build from the dermoscopic attribute and segmentation information, gradually adding information until the samples contain all image's pixels plus an additional channel containing extra dermoscopic attribute and segmentation information.  }\label{fig:isicmodified}
\end{figure*}

We introduce further modifications that are only possible with the dermoscopic attributes masks available on the ISIC dataset. Please refer to examples in Figure~\ref{fig:isicmodified}.

\textbf{Grayscale Attributes}: To compose each image in this set, we use a lesion's masks from ISIC that show the location of five dermoscopic attributes and the same lesions' segmentation masks. The skin without lesion, the lesion without markers, and each dermoscopic attribute are assigned a different value, equally spaced from each other. Dermatologists look for this information to diagnose skin lesions, and it is the basis for different medical algorithms, therefore being one of the most critical parts of the image.

\textbf{RGB Attributes}: This dataset only shows the RGB values of the regions of the image that belongs to an annotated dermoscopic attribute, and mask the others. This way, the network does not know in principle what are the skin patterns in the image or how many of them are present, but it gain access to their RGB values. We keep the segmentation mask information from \textit{Grayscale Attributes} in this set to display some information for cases that do not present any skin patterns. 
ISIC's annotation over the dermoscopic attributes is not as detailed as Atlas'. By letting the network analyze the RGB pixels that belong to a dermoscopic attribute, we are forcing the network to focus on the attributes, to discover more details about them (\eg typical or atypical, regular or irregular, etc.), and to rely the classification on this information.

\textbf{Traditional$+$Grayscale Attributes}: Here we aim to guide the learning process by giving to the network extra information that is very relevant to dermatologists. We concatenate a fourth channel to the \textit{Traditional} image, containing the information described in the \textit{Grayscale Attributes}. We need to adapt the network to receive the extra channel in the input. We add an extra convolutional layer at the beginning of the network, initialized to prioritize receiving information from the RGB channels, and progressively learn to make use of the mask provided. We expected the results to be better than \textit{Traditional} since we are adding clinically-meaningful information to guide the network to a better understanding of the process according to human knowledge.

\subsection{Results and Discussion}

\begin{figure}[th]
    \centering
    \includegraphics[trim={0 0.55cm 0 1.12cm},clip,width=1\linewidth]{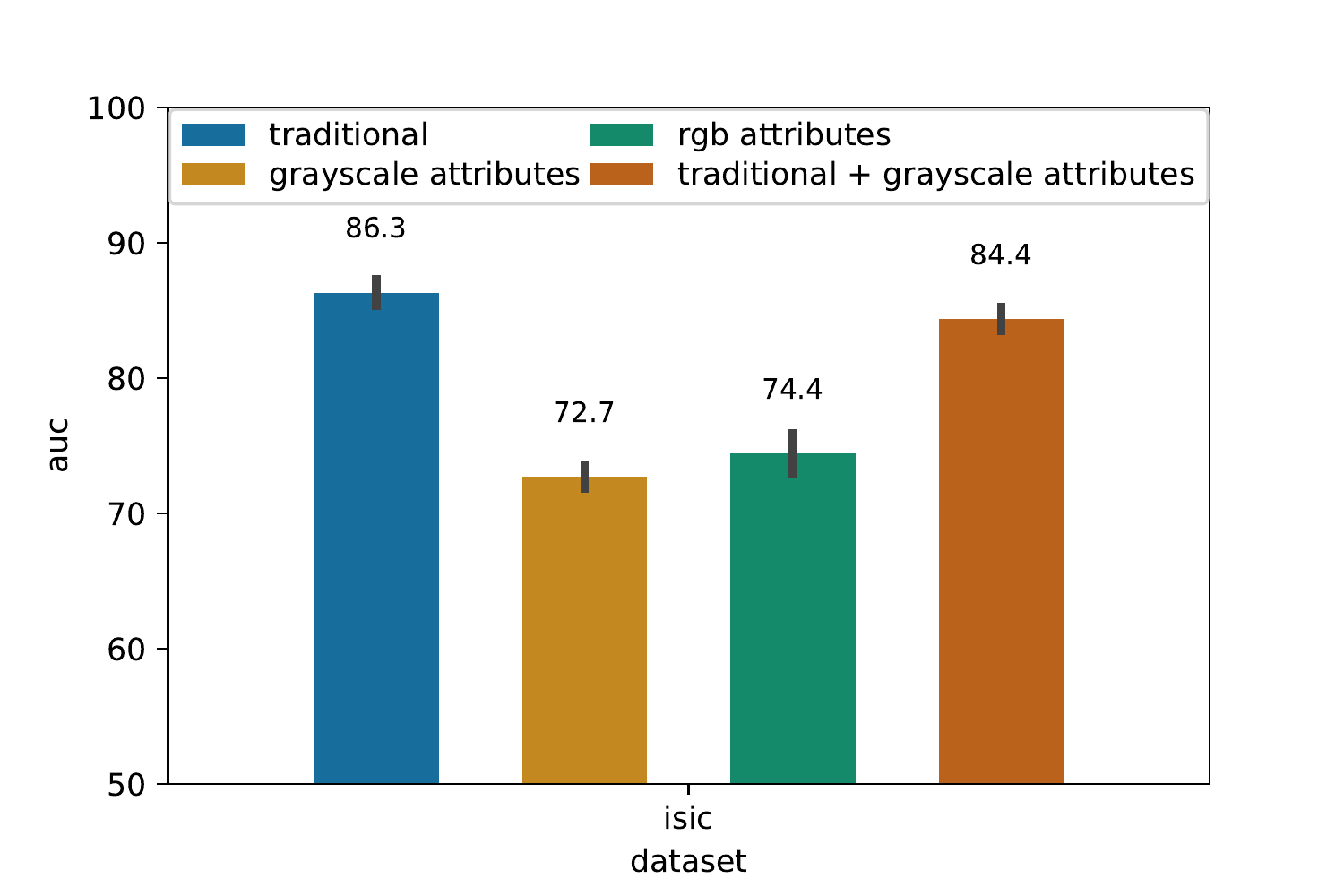}
    \caption{Performance comparison of the different sets of images with the ISIC dataset. Surprisingly, when we try to simplify the learning process, feeding the network with dermoscopic attributes that are clinically-meaningful, the result does not improve.\vspace{-0.25cm}}
    \label{fig:constructionisic}
\end{figure}

We show in Figure \ref{fig:constructionisic} our results evaluating all different sets on the ISIC dataset. 

Our attempt to guide the network's learning process did not yield better results. 
Starting from \textit{Grayscale Attributes}, we are feeding the network with enough information to verify global patterns present in the lesion, and location of some local features (pigment network, globules, streaks, negative network, and milia-like cysts). We note that the dermoscopic attributes information is not as detailed as the one present in Atlas, and this may affect the capability of the network to make correct predictions.

In \textit{RGB Attributes}, we add pixel information to the images. That enables the network to learn details about each different dermoscopic attribute and improve classification. However, we did not observe that behavior. The extra information did not help the network to improve its understanding of the problem.

In \textit{Traditional$+$Grayscale Attributes}, where we are add\-ing clinical relevant information to the usual classification procedure to guide the learning process, the result did not improve as well in comparison to the \textit{Traditional} baseline. 

\section{Conclusion}\label{sec:conclusion}

If we hide the same lesion information from the networks, can it still learn patterns that help differentiate benign from malignant lesions? We believe that when a~model learns to classify malignant lesions by analyzing only the skin ---without information on the borders, biological markers or lesions' diameter--- it strongly relies on patterns introduced during image acquisition and general dataset bias.

Surprisingly, the result when feeding the network with clinically-meaningful information from the dermoscopic attribute maps (\textit{Grayscale Attributes} and \textit{RGB Attributes} sets) is worse than feeding it only with healthy skin information (\textit{Only Skin} and \textit{Bounding Box} sets). That leads us to believe that also our networks' results towards both datasets is optimistic, not only the performance of 7-points over Atlas (which is expected).

That problem is critical for deploying automated skin lesion analysis. When performing in the real world, we want the network to be as unbiased as possible to make decisions based on clinical features. Therefore, it is urgent to understand the current bias in the datasets used to train and evaluate our works.\vspace{-0.1cm}

\section*{Acknowledgments}
\vspace{-0.1cm}
A. Bissoto and S. Avila are partially funded Google LARA 2018. A. Bissoto is also partially funded by CNPq. E. Valle is partially funded by a CNPq PQ-2 grant (311905/2017-0). This work was funded by grants from CNPq (424958/2016-3), FAPESP (2017/16246-0) and FAEPEX (3125/17). The RECOD Lab receives addition funds from FAPESP, CNPq, and CAPES. We gratefully acknowledge NVIDIA for the donation of GPU hardware.

{\small
\bibliographystyle{ieee_fullname}
\bibliography{egbib}
}

\end{document}